%% file: main.tex
\documentclass[11pt,a4paper]{article}
\usepackage[hyperref]{eacl2021}
\usepackage{times}
\usepackage{latexsym}

\usepackage{microtype}

\aclfinalcopy 

\usepackage{amsmath}
\DeclareMathOperator*{\argmax}{arg\,max}

\usepackage{subcaption}
\usepackage{graphicx}
\usepackage{comment}
\usepackage{booktabs}

\newcommand{\modelname}[1]{\textsc{ARedSum}}
\newcommand{\modelnameSEQ}[1]{\textsc{ARedSum-Seq}}
\newcommand{\modelnameCTX}[1]{\textsc{ARedSum-Ctx}}

\title{\modelname{}: Adaptive Redundancy-Aware Iterative Sentence Ranking for Extractive Document Summarization}

\author{Keping Bi$^{1}$, Rahul Jha$^2$, W. Bruce Croft$^1$, Asli Celikyilmaz$^3$ \\
  $^1$ University of Massachusetts Amherst, $^2$ Microsoft, $^3$ Microsoft Research AI \\
  \texttt{\{kbi, croft\}@cs.umass.edu, rajh@microsoft.com, asli@ieee.org}
}

\date{}

\begin{document}
\maketitle

\begin{abstract}

Redundancy-aware extractive summarization systems score the \textit{redundancy} of the sentences to be included in a summary either jointly with their \textit{salience} information or separately as an additional sentence scoring step.
Previous work shows the efficacy of jointly scoring and selecting sentences with neural sequence generation models. It is, however, not well-understood if the gain is due to better encoding techniques or better redundancy reduction approaches. 
Similarly, the contribution of salience versus diversity components on the created summary is not studied well.  
Building on the state-of-the-art encoding methods for summarization, we present two adaptive learning models: \modelnameSEQ{} that jointly considers salience and novelty during sentence selection; and a two-step \modelnameCTX{} that scores salience first, then learns to balance salience and redundancy, enabling the measurement of the impact of each aspect. 
Empirical results on CNN/DailyMail and NYT50 datasets show that by modeling diversity explicitly in a separate step, \modelnameCTX{} achieves significantly better performance than \modelnameSEQ{} as well as state-of-the-art extractive summarization baselines.

\end{abstract}

\input{outline.tex}
\input{introduction.tex}
\input{related-work.tex}
\input{iterative-sentence-ranking.tex}

\input{redundancy-aware-summ.tex}
\input{experimental-setup.tex}

\input{experimental-results.tex}

\input{conclusion.tex}

\bibliography{references}
\bibliographystyle{acl_natbib}


\end{document}

%% file: introduction.tex
\section{Introduction}
\label{sec:introduction}
Extractive summarization is the task of creating a summary by identifying and concatenating the most important sentences in a document \cite{liu2019text, zhang-etal-2019-hibert, zhou2018neural}. 
Given a partial summary, the decision to include another sentence in the summary depends on two aspects: \textit{salience}, which represents how much information the sentence carries; and \textit{redundancy}, which represents how much information in the sentence is already included in the previously selected sentences.

Although there have been a few studies on redundancy a long time ago, most recent research on extractive summarization focuses on salience alone. They usually model sentence salience as a sequence labeling task \cite{kedzie2018content, cheng2016neural} or classification task \cite{ zhang-etal-2019-hibert} and do not conduct redundancy removal. 
Previous methods that consider redundancy usually use a separate step after salience scoring to handle redundancy, denoted as sentence selection \cite{carbonell1998use,mcdonald2007study,lin2011class}. Sentence selection often follows a greedy iterative ranking process that outputs one sentence at a time by taking into account the redundancy of candidate sentences with previously selected sentences.

Several approaches for modeling redundancy in sentence selection have been explored: heuristics-based methods such as Maximal Marginal Relevance (MMR) \cite{carbonell1998use}, Trigram Blocking (\textsc{TriBlk}) \cite{liu2019text}, or model based approaches \cite{ren-etal-2016-redundancy}, etc.
Heuristic-based methods are not adaptive since they usually apply the same rule to all the documents. Model-based approaches depend heavily on feature engineering and learn to score sentences via regression with point-wise loss, which has been shown to be inferior to pairwise loss or list-wise loss in ranking problems \cite{liu2009learning}. 

Redundancy has also been handled jointly with salience during the scoring process using neural sequence models \cite{zhou2018neural}. \textsc{NeuSum} \cite{zhou2018neural} scores sentences considering their salience as well as previous sentences in the output sequence and learns to predict the sentence with maximum relative gain given the partial output summary. 
Despite the improved efficacy, it is not well-understood if the gain is due to better encoding or better redundancy-aware iterative ranking approaches (i.e., the sequence generation). 

In this work, we propose to study different types of redundancy-aware iterative ranking techniques for extractive summarization that handle redundancy separately or jointly with salience. Extending \textsc{BertSumExt} \citep{liu2019text}, a state-of-the-art extractive summarization model, which uses heuristic-based Trigram Blocking (\textsc{TriBlk}) for redundancy elimination, we propose two supervised redundancy-aware iterative sentence ranking methods for summary prediction.  
Our first model, \modelnameSEQ{}, introduces a transformer-based conditional sentence order generator network to score and select sentences by jointly considering their salience and diversity within the selected summary sentences.  
Our second model, \modelnameCTX{}, uses an additional sentence selection model to learn to \textit{balance} the salience and redundancy of constructed summaries. It incorporates surface features (such as n-gram overlap ratio and semantic match scores) to instrument the diversity aspect.
We compare the performance of our redundancy-aware sentence ranking methods with trigram-blocking \cite{liu2019text}, as well as summarization baselines with or without considering redundancy on two commonly used datasets, CNN/DailyMail and New York Times (NYT50).
Experimental results show that our proposed \modelnameCTX{} can achieve better performance by reducing redundancy and outperform all the baselines on these two datasets. The model's advantage can be attributed to its adaptiveness to scenarios in which redundancy removal has different potential gains.

In summary, our contributions are:
1) we propose two redundancy-aware iterative ranking methods for extractive summarization extending \textsc{BertSumExt};
2) we conduct comparative studies between our redundancy-aware models as well as the heuristic-based method that \textsc{BertSumExt} uses; 
3) our proposed \modelnameCTX{} significantly outperforms \textsc{BertSumExt} and other competitive baselines on CNN/DailyMail and NYT50. 

%% file: related-work.tex
\section{Related Work}
\label{sec:related_work}

Extractive summarization methods are usually decomposed into two subtasks, i.e., sentence scoring and sentence selection, which deal with salience and redundancy, respectively. 

\paragraph{\textbf{Salience Scoring.}} Graph-based models are widely used methods to score sentence salience in summarization \cite{erkan2004lexrank,mihalcea2004textrank,wan2006improved}. 
There are also extensions to such methods, e.g., with clustering \cite{wan2008multi} or leveraging graph neural networks \cite{wang2020heterogeneous}.
Classical supervised extractive summarization uses classification or sequence labeling methods such as Naive Bayes \cite{kupiec1999trainable}, maximum entropy \cite{osborne2002using}, conditional random fields \cite{galley2006skip} or hidden markov model \cite{conroy2004left}. Human engineered features are heavily used in these methods such as word frequency and sentence length \cite{nenkova2006compositional}. 

In recent years, 
neural models have replaced older models to score the salience of sentences. 
Hierarchical LSTMs and CNNs have replaced manually engineered features. LSTM decoders are employed to do sequence labeling \cite{cheng2016neural, nallapati2017summarunner,kedzie2018content}. These architectures are widely used and also extended with reinforcement learning \cite{narayan2018ranking, dong2018banditsum}. 
More recently, summarization methods based on \textsc{Bert} \cite{devlin2018bert} have been shown to achieve state-of-the-art performance \cite{liu2019text, zhang-etal-2019-hibert,zhong2019searching,zhou2020level} on salience for extractive summarization. 

\paragraph{\textbf{Sentence Selection.}}
There are relatively fewer methods that study sentence selection to avoid redundancy. Integer Linear Programming based methods \cite{mcdonald2007study} formulate sentence selection as an optimizing problem under the summary length constraint. \citet{lin2011class} propose to find the optimal subset of sentences with submodular functions. 
Greedy strategies such as Maximal Marginal Relevance (MMR) \cite{carbonell1998use} 
select the sentence that has maximal salience score and is minimally redundant iteratively. Trigram blocking \cite{liu2019text} follows the intuition of MMR and filters out sentences that have trigram overlap with previously extracted sentences. 
\citet{ren-etal-2016-redundancy} leverage two groups of handcrafted features to capture informativeness and redundancy respectively during sentence selection. 
In contrast to learning a separate model for sentence selection, \citet{zhou2018neural}  propose to jointly learn to score and select sentences with a sequence generation model. However, it is not compared with other redundancy-aware techniques, and it is not clear whether its improvement upon other methods is from the sequence generation method or the encoding technique. 

In this paper, we compare the efficacy of different sentence selection techniques grounded on the same \textsc{Bert}-based encoder. 
We propose two models that either conduct redundancy removal with a separate model or jointly with salience scoring and compare them with a heuristic-based method. As far as we know, our work is the first to conduct comparative studies on different types of redundancy-aware extractive summarization methods.

%% file: iterative-sentence-ranking.tex
\section{Iterative Sentence Ranking} 
\label{sec:problem_formulation}
We formulate single document extractive summarization as a task of iterative sentence ranking.
Given a document $D=\{s_1, s_2, \cdots, s_{L}\}$ of $L$ sentences, the goal is to extract $t$ sentences, i.e., $\hat{S}_t=\{\hat{s}_k|1\leq k \leq t, \hat{s}_k \in D\}$, from $D$ that can best summarize it. 
With a limit of selected sentence count $l$, the process of extracting sentences can be considered as a $l$-step iterative ranking problem. At each $k$-th step ($1 \leq k \leq l$), given the current summary $\hat{S}_{k-1}$, a new sentence $s_k$ is selected from the remaining sentences $D \setminus \hat{S}_{k-1}$ and added to the summary. Function $M(\hat{S}_k;S^*)$\footnote{In $\S$~\ref{sec:methods} and experiments, we use \textsc{Rouge} to define $M(\cdot)$} measures the similarity between the extracted summary $\hat{S}_{k}$ and the ground truth summary $S^*$. The objective is to learn a scoring function $f(\cdot)$ so that the best sentence $\hat{s}_k$ selected according to $f(\cdot)$ can maximize the gain of the output summary:
\begin{align}
\begin{split}
&\argmax_{f} M(\{\hat{s}_k\} \cup \hat{S}_{k-1});S^*) \\
&\hat{s}_k = \argmax_{s_i\in D \setminus \hat{S}_{k-1}} f(\{s_i\} \cup \hat{S}_{k-1})
\end{split}
\end{align}
$\hat{s}_k$ needs to be both salient in the document and novel in the current context $\hat{S}_{k-1}$. Note that at the beginning $\hat{S}_0 = \emptyset$. 

Since ground truth summaries $S^*$ of existing summarization corpora are usually abstractive summaries written by experts, previous studies on extractive summarization usually extract a group of pseudo ground truth sentences $\hat{S}^*$ from $D$ based on their similarities to the ground truth summaries $S^*$ for training purposes. Then labels 1 and 0 are assigned to sentences in $\hat{S}^*$ and the other sentences in $D$. In this case, $M(\hat{S}_t;\hat{S}^*)$ is used to guide training instead of $M(\hat{S}_t;S^*)$.

%% file: redundancy-aware-summ.tex
\begin{figure*}
    \centering
    \includegraphics[width=0.9\textwidth]{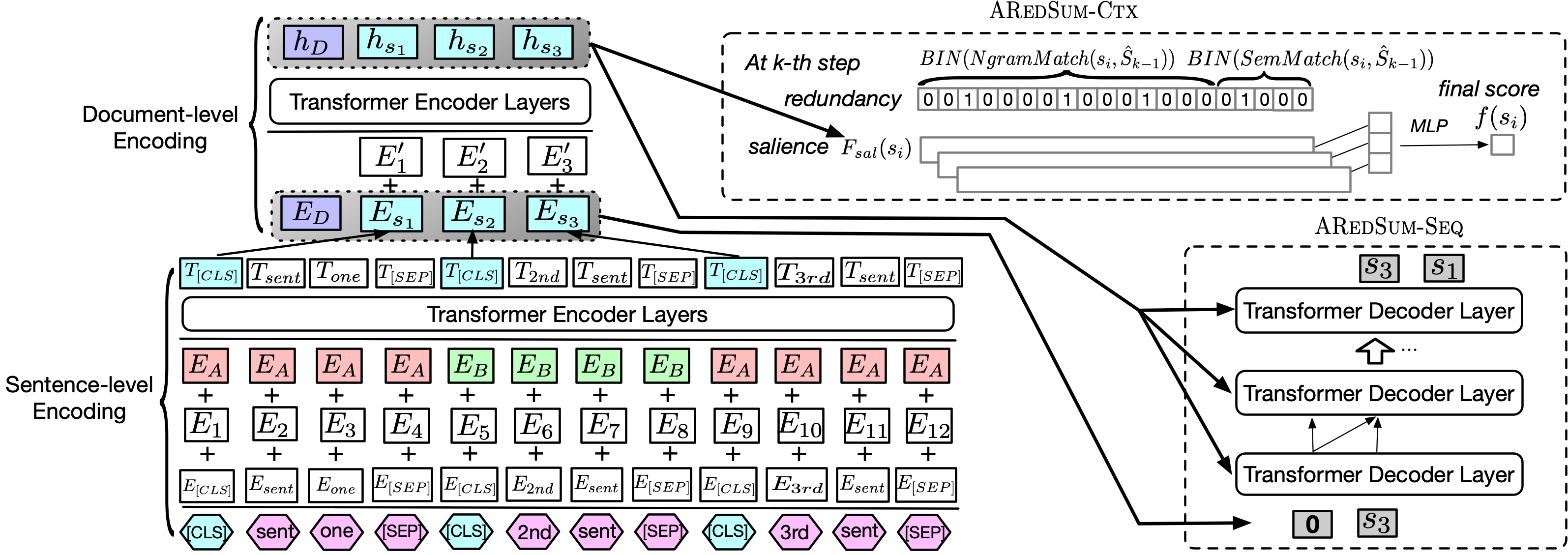}
    \caption{Overview of the proposed models \modelnameSEQ{} and \modelnameCTX{} sharing the same \textsc{Bert}-based encoder from \textsc{BertSumExt}.}  
\label{fig:model}
\end{figure*}

\section{Redundancy-Aware Summarization}
\label{sec:methods}
Most recent redundancy-aware extractive summarization systems use heuristics to select diverse sentences after salience scoring \cite{cao2015learning, ren2017leveraging, liu2019text}. Among them, \textsc{BertSumExt} \cite{liu2019text} is a state-of-the-art model with trigram-blocking (\textsc{TriBlk}) that reduces redundancy by filtering out sentences that have trigram overlap with previously selected ones at each time step. As we empirically show later in $\S$~\ref{sec:results}, heuristics can be effective on some datasets yet harmful on others since it applies the same rule to all the documents.

In contrast, we present an adaptive learning process for redundancy-aware extractive summarization, \modelname{}, and introduce two methods, \modelnameSEQ{} and \modelnameCTX{}, extending \textsc{BertSumExt} by either consider redundancy jointly with salience during sentence scoring or separately with an additional selection model. 

\subsection{Document Encoder}
\label{subsec:doc_encoder}
First, we introduce the sentence and document encoder shared by both our variations of \modelname{}, shown in Figure \ref{fig:model}.
In sentence-level encoding, a [SEP] token is appended to each sentence to indicate the sentence boundaries and a [CLS] token is inserted before each sentence in the document to aggregate the information of the sentence. 
In addition to token and positional embeddings, as in \textsc{BertSumExt} \cite{liu2019text}, we also use interval segment embeddings $E_A$ and $E_B$ to distinguish sentences at odd and even positions in the document respectively. Following multiple transformer encoder layers, we represent each sentence $s_i$ by the output representation of the [CLS] symbol preceding $s_i$. These symbols capture the features of the following tokens in the sentence while attending over all other tokens in the document through the transformer layers.

We further conduct document-level encoding on the sentence-level representations from the [CLS] tokens, denoted as $E_{s_i}$, as well as their positional embeddings, $E^{\prime}_i$, with another stack of transformer layers. We add a document embedding $E_D$ before the sequence of sentence embeddings to represent the whole document. The final representation of $D$ and each sentence $s_i$ can be obtained from the output of the multiple transformer layers, denoted as $h_{s_i}$ and $h_D$. 


\subsection{\modelnameSEQ{}: Sequence Generation}
\label{subsec:seq_gen_model}
Our first model, \modelnameSEQ{}, strictly considers the \textit{order} of the target selected sentences while jointly modeling the redundancy and salience of the next sentence. It uses a transformer decoder module \cite{vaswani2017attention} to learn to select and order a sequence of sentences from the document as a summary. Our model is different from standard auto-regressive decoders. Each decoder block takes in a sequence of tokens (word-units) as input to generate the next possible token from a pre-defined vocabulary. Instead, our decoder is a conditional model that takes a sequence of sentence representations as input and selects a sentence with the maximum gain to be included in the summary from the rest of the document's sentences.

Following a standard transformer encoder-decoder architecture  \cite{vaswani2017attention}, at each decoding step $k$, 
a current hidden state is obtained with a stack of transformer decoder layers:
\begin{equation}
\begin{split}
\!\!h'_{\hat{s}_{k-1}}\!\!\!\!\! =\! \textsc{Dec}([E_{\hat{s}_1}\!, \!\cdots\!,\!E_{\hat{s}_{k-1}}],\! [h_{s_1},\!h_{s_2}\!,\!\cdots\!,\!h_{s_L}]) \!\!\!
\end{split}
\end{equation}
where $[h_{s_1}, h_{s_2}, \cdots, h_{s_L}]$ are the output sentence representations after the document-level encoding in Figure \ref{fig:model} and $[E_{\hat{s}_1}, \cdots, E_{\hat{s}_{k-1}}]$ are the embeddings of the so far selected sentences. $\hat{S}_{k-1} = \emptyset$ and $E_{\hat{s}_0} = \boldsymbol{0}$ when $k=1$. Note that sentence embeddings that are fed to the document-level transformer encoders, i.e., $E_{s_1}, \cdots, E_{s_L}$, are used to represent the sentences in the target decoding space. 
Then, a two-layer MLP is used to score a candidate sentence $s_i$ given the hidden state $h'_{\hat{s}_{k-1}}$:
\begin{equation}
    o_l(s_i) = W_{2s} \tanh (W_{1s} [h'_{\hat{s}_{k-1}}; E_{s_i}]))
\end{equation}
where $W_{2s}$ and $W_{1s}$ are the weights of the MLP (we omit the bias parameters for simplicity), and $[;]$ denotes vector concatenation. In case the salience of $s_i$ is not sufficiently captured in $o_l(s_i)$, we calculate a matching score $o_g(s_i)$ between $s_i$ and the global context $D$, regardless of which sentences are selected previously
\footnote{Emphasizing salience with $o_g$ enhances the performance. }: 
\begin{equation}
\label{eq:o2}
    o_g(s_i) = \tanh (h_D W_{ds}h_{s_i})
\end{equation}
where $W_{ds}$ is matrix for bilinear matching; 
$h_D$ and $h_{s_i}$ are the embeddings of the document $D$ and sentence ~$s_i$ output by the document-level encoder. 
The final score is the linear combination of $o_l$ and $o_g$ using the weight $W_o$:
\begin{equation}
    o(s_i) = W_o (o_l(s_i) + o_g(s_i))
\end{equation}
The probability of any sentence $s_i$ being selected at the $k$-th step is the softmax of $o(s_i)$ over the remaining candidate sentences $s_j$ in $D$:
\begin{equation}
\label{eq:seq_o_prob}
P(\hat{s}_k\!\! = \!\! s_i| \hat{S}_{k-1}) = \frac{\exp({o(s_i))}}{\sum_{s_j \in D \setminus \hat{S}_{k-1}} \!\! \exp(o(s_j))}
\end{equation}

Following \textsc{NeuSum} \cite{zhou2018neural}, we train \modelnameSEQ{} to optimize for the \textit{relative} \textsc{rouge-f1} gain of each sentence with respect to so-far selected sentences $\hat{S}_{k-1}$. 
\begin{equation}
    g(s_i) \!= \!M(\{s_i\} \! \cup \! \hat{S}_{k-1};S^*) \!-\! M(\hat{S}_{k-1};S^*)
\end{equation}
where $M(\{s_i\} \! \cup \! \hat{S}_{k-1};S^*)$ and $M(\hat{S}_{k-1};S^*)$ measure the \textsc{Rouge-F1} between the golden summary $S^*$ and the so-far selected sentences $\hat{S}_{k-1}$ with and without the candidate $s_i$ respectively.
We rescale the gain $g(s_i)$ to [0,1] in case of negative values using a min-max normalization and get $\tilde{g}$. Then we use a softmax function with a temperature $\tau$ on the rescaled gain to produce a target distribution:
\begin{equation}
    \label{eq:target_rouge}
    Q(s_i) = \frac{\exp(\tau \tilde{g}(s_i))}{\sum_{s_j \in D \setminus \hat{S}_{k-1}} \!\! \exp(\tau \tilde{g}(s_j))}
\end{equation}
The final training objective is to minimize the KL divergence between the probability distribution of sentence scores (Eq.~\ref{eq:seq_o_prob}) and their relative rouge gain (Eq.~\ref{eq:target_rouge}), i.e., $KL(P(\cdot)||Q(\cdot))$. This objective can be considered as a listwise ranking loss \cite{ai2018learning} that maximizes the probability of the target sentence while pushing down the probabilities of the other sentences. 
In this way, \modelnameSEQ{} combined the sentence scoring and selection in the same decoder framework, and the redundancy is implicitly captured by optimizing the \textsc{rouge} gain. 

\subsection{\modelnameCTX{}: Context-aware Sentence Ranker}
\label{subsec:contextaware_ranker}
We introduce a second model, \modelnameCTX{}, a context-aware ranker that scores salience first and then selects a sentence according to both its salience and redundancy adhering to the previously extracted sentences as context, as shown in Figure~\ref{fig:model}.
In \modelnameCTX{}, we use a two-step process for scoring and selecting sentences for learning to construct a summary: In the \textit{salience ranking} step, we focus on learning the salience of the sentences, while in the \textit{ranking for sentence selection} step, we represent the redundancy explicitly via surface features and use a ranker to decide to promote or demote sentences based on their scores given the joint degree of their salience and redundancy. 

\paragraph{\textbf{Salience Ranking}.}
By assuming that the sentence salience is independent of the previously selected sentences, we design the salience ranking of \modelnameCTX{} as a single step process rather than an iterative one. 
We measure the probability of a sentence to be included in $\hat{S}^*$ using a scoring function $F_{\text{sal}}$ based on the bilinear matching between $h_D$ and $h_{s_i}$, the transformer output after the document-level encoding, same as in Eq. \ref{eq:o2}. 
\begin{equation}
    \label{eq:salience}
    F_{\text{sal}}(s_i) = \frac{\exp{h_{D} W_{ds} h_{s_i}}}{\sum_{j=1}^{j=L} \exp{h_{D} W_{ds} h_{s_j}}}
\end{equation}
The learning objective is to maximize the log likelihood of the summary sentences in the training data:
\begin{equation}
    \mathcal{L} = \sum_{s_i \in \hat{S}^*}\log F_{\text{sal}}(s_i)
\end{equation}

\paragraph{\textbf{Redundancy Features}.}
In the selection step, we represent redundancy explicitly to let the model focus on learning how to balance salience and redundancy. 
We extract ngram-matching and semantic-matching features at each $k$-th step to indicate the redundancy of a candidate sentence $s_i$ given the so-far selected sentences, i.e., $\hat{S}_{k-1}$.
The ngram-matching feature $f_{n\text{-gram}}$ is computed as:
\begin{equation}
    f_{n\text{-gram}} \!= \! \frac{|n\text{-gram}(\hat{S}_{k-1}) \cap n\text{-gram}(s_i)|}{n\text{-gram}(s_i)}
\end{equation}
where $n$-gram$(x)$ is the set of $n$ contiguous words in $x$. We collect $f_{n\text{-gram}}$ for $n\!=\!\{1,2,3\}$. 
We also compute the semantic-matching feature $f_{sem}$:
\begin{equation}
    f_{\text{sem}} = \max_{\hat{s}_j \in \hat{S}_{k-1}} \cos(h_{s_i}, h_{\hat{s}_j})
\end{equation}
Since most cosine values between output embeddings from the transformer layers fall in a small range near to 1, we apply a min-max normalization on $f_{\text{sem}}$ to enlarge the value differences and obtain a updated feature $\tilde{f}_{\text{sem}}$. 

The impact of redundancy features on final scores is not linear. Sentences with high redundancy values should be punished more. 
To capture the effect of the redundancy features at different value sections, 
we equally divide the range of $[0,1]$ to $m$ bins and discretize each feature to the corresponding bin according to its value,
as shown in Figure~\ref{fig:model}. In this way, we convert each feature into a one-hot vector of length $m$ and then we concatenate them to obtain a overall redundancy feature vector $F_{\text{red}}(s_i)=[f'_{1\text{-gram}};f'_{2\text{-gram}};f'_{3\text{-gram}};\tilde{f}'_{\text{sem}}]$ where $f'$ represents the one-hot vector after binning $f$.  

\paragraph{\textbf{Ranker for Sentence Selection}.}
In the sentence selection step, \modelnameCTX{} only needs to learn how to score a sentence based on its redundancy features $F_{\text{red}}(s_i)$ and its salience score $F_{\text{sal}}(s_i)$ from Eq. \ref{eq:salience}. Note that the first selected sentence is the one ranked with the largest salience score. 
We use a three-dimensional matrix $W_{\mathcal{F}}$ to do a bilinear matching between the redundancy features and salience score and obtain a output matching vector with dimension $d$. Then we apply a single-layer MLP on top to output a final score:
\begin{equation}
\label{eq:contextr_score}
f(s_i) = W_f\tanh(F_{\text{sal}}(s_i)W_{\mathcal{F}}F_{\text{red}}(s_i))
\end{equation}
During training, we randomly select $1, 2, \cdots, l\text{-1}$ sentences from the extracted ground-truth set $\hat{S}^*$ as the context and let the model learn to find the next sentence that is both salient and novel, where $l$ is the maximum number of sentences to be included in the predicted summary.
The training objective is the same as in $\S$~\ref{subsec:seq_gen_model} except that $o(s_i)$ in Eq. \ref{eq:seq_o_prob} is replaced with $f(s_i)$ in Eq. \ref{eq:contextr_score}. 
In contrast to \modelnameSEQ{} where the target output is an ordered sequence, the loss of \modelnameCTX{} is not order-sensitive since the goal is always to predict the next best sentence given a set of unordered selected sentences as context. 


%% file: experimental-setup.tex
\section{Experimental Setup}
\label{sec:exp_setup}
\subsection{Datasets}
We evaluate our model on two standard extractive summarization datasets, namely CNN/DailyMail \cite{hermann2015teaching} and NewYork Times (NYT) \cite{sandhaus2008new}. 

\textbf{CNN/DailyMail} contains news articles associated with a few bullet points as the article's highlight. We use the standard splits of \citet{hermann2015teaching} which has 287,226 documents for training, 13,368 for validation, and 11,490 for testing. We conduct preprocessing following the same method in \citet{liu2019text}. Entities are \textit{not anonymized} in our experiments as in \citet{zhou2018neural, see2017get, zhang-etal-2019-hibert, liu2019text}. We truncate articles up to 512 tokens.
To collect sentence labels for extractive summarization, we use a greedy strategy similar to \cite{nallapati2017summarunner,zhang-etal-2019-hibert}. We label the subset of sentences that can maximize ROUGE scores against the human-generated summary as 1 (sentence to be included in the summary). The remaining ones are labeled as 0. 

\textbf{NYT50} is an annotated corpus of the New York Times. 
Following \citet{paulus2017deep} and \citet{durrett2016learning}, we discard marks and words such as ``(s)'' and ``photo'' at the end of the abstract and filter out the articles with summaries shorter than 50. 
We sort the articles chronologically and split the data into training/validation/test sets according to the ratio of $0.8/0.1/0.1$, yielding 133,602/16,700/16,700 documents, respectively. We following the same remaining steps for preprocessing and extractive label collection as the CNN/DailyMail. 

\subsection{Implementation Details}
Our implementation
\footnote{\url{https://github.com/kepingbi/ARedSumSentRank}} 
is based on PyTorch and \textsc{BertSum}\cite{liu2019text} \footnote{\url{https://github.com/nlpyang/BertSum}}. We use ``bert-base-uncased'' version of 
\textsc{Bert}\footnote{\url{https://git.io/fhbJQ}} to do sentence-level encoding. 
We fine-tune our models using the objective functions in $\S$~\ref{sec:methods}. 
We set the number of document-level transformer layers to 2. The dropout rate in all layers is 0.1.
We search the best value of $\tau$ in Eq. \ref{eq:target_rouge} in $\{10,20,40,60\}$. 
We train our models using the Adam optimizer with $\beta_1=0.9$, and $\beta_2 = 0.999$ for 2 epochs. We schedule the learning rate according to \citet{vaswani2017attention} with initial value $2e\text{-3}$ and 10,000 warm-up steps.

We use teacher-forcing to train \modelnameSEQ{}. To learn the $k$-th sentence in the target sequence, we replace the first $k-1$ input sentences with other random sentences in the document with the probability of 0.2 \footnote{We found this number after light parameter value sweep.}. We hypothesize that if the previously selected sentence is not always the golden (right) sentence, it can improve the model's robustness during training.
We use two transformer layers in the decoder. 

For \modelnameCTX{}, 
we train the salience ranker using the same settings as in \citet{liu2019text}, and all the parameters in the salience ranker are fixed when we train the ranker for selection. This ensures that the salience score of each sentence stays the same during sentence selection. 
We select the optimum size of the bins for discretized redundancy features by sweeping the values in $\{10,20,30\}$ and the size of the output dimension $d$ of $W_\mathcal{F}$ in Eq. \ref{eq:contextr_score} using $\{5,10,20,30\}$.

\subsection{Baselines}
We compare our models to the state-of-the-art extractive summarization model \textsc{BertSumExt} \cite{liu2019text}, which uses Trigram Blocking (\textsc{TriBlk}) \cite{paulus2017deep} to filter out sentences with trigram overlap with previously extracted sentences. 
We report the performance of \textsc{BertSumExt} with and without \textsc{TriBlk} separately to show the impact of this heuristic. 

We also compare against other baselines including: \textsc{Lead3}, NN-SE \cite{cheng2016neural}, \textsc{SummaRuNNer} \cite{nallapati2017summarunner}, Seq2Seq \cite{kedzie2018content}, \textsc{NeuSum} \cite{zhou2018neural}, and \textsc{HiBert} \cite{zhang-etal-2019-hibert} along with \textsc{Oracle} for upper bound on performance. 
\textsc{Lead3} is a commonly used effective baseline that extracts the first 3 sentences in the document.
NN-SE \cite{cheng2016neural} and \textsc{SummaRuNNer} \cite{nallapati2017summarunner} both formulate extractive summarization as a sequence labelling task. NN-SE uses unidirectional GRU for both the encoding and decoding processes. 
\textsc{SummaRuNNer} encodes sentences with BiGRU and considers salience, redundancy, absolute and relative positions of sentences during scoring.
Seq2Seq \cite{kedzie2018content} conducts binary classification by encoding the sentences with a bidirectional GRU (BiGRU) and using a separate decoder BiGRU to transform each sentence as a query vector that attends to the encoder output. 
\textsc{NeuSum} \cite{zhou2018neural} learns to jointly score and select sentences using a sequence-to-sequence model to optimize the marginal \textsc{Rouge} gain and reduce redundancy implicitly. 
\textsc{HiBert} \cite{zhang-etal-2019-hibert} pre-trains a hierarchical BERT for extractive summarization without dealing with redundancy.
Among these methods, \textsc{NeuSum} implicitly reduces redundancy by jointly scoring and selecting sentences with a sequence generation model. \textsc{SummaRuNNer} considers redundancy during the sequence labeling. The other baselines do not conduct redundancy removal.

%% file: experimental-results.tex
\section{Results and Discussion}
\label{sec:results}
\subsection{Automatic Evaluation Results}
\label{subsec:overall_perf}
Following earlier work \cite{zhou2018neural,liu2019text}, we include 3 sentences as the summaries for each system for a fair comparison. 
We evaluate the full-length \textsc{Rouge-f1} \cite{lin2004rouge} of the extracted summaries and report \textsc{Rouge-1}, \textsc{Rouge-2} and \textsc{Rouge-L} which indicates the unigrams, bigrams overlap and longest common subsequence against human edited summaries.
The full-length \textsc{Rouge-f1} \cite{lin2004rouge} scores of the extracted summaries are evaluated using the official Perl script\footnote{\url{https://github.com/andersjo/pyrouge/tree/master/tools/ROUGE-1.5.5}} for both CNN/DailyMail and NYT50.
The results of \textsc{NeuSum} and \textsc{HiBert} are taken from their original papers while we obtained the rest of the results by re-running the models. 
Since in the previous work~\cite{liu2019text,zhang-etal-2019-hibert,paulus2017deep} there are no consistent ways of pre-processing the NYT dataset for extractive summarization, we only report the evaluation results from the models we re-trained on this dataset in Table \ref{tab:nyt_result}.

\textbf{CNN/DailyMail}.
Results shown in Table \ref{tab:cnndm_result} are all comparable as we use the same \textit{non-anonymized} version of CNN/DailyMail.
\begin{table}
    \caption{Full-length \textsc{Rouge} (RG) F1 evaluation (\%) on the CNN/DailyMail test set. a/b and A/B indicate significant improvements over \textsc{BertSumExt}/\textsc{BertSumExt}+\textsc{TriBlk} with $p<0.05$ and $p<0.0001$ respectively.} 
    \centering
    \label{tab:cnndm_result}  
    \small
    \begin{tabular}{l  p{0.9cm} p{0.9cm}  p{0.9cm} }
    \toprule
    Model & RG-1 & RG-2 & RG-L \\
    \toprule
    \textsc{Oracle} & 52.59 & 31.24 & 48.87 \\
    \textsc{Oracle}+\textsc{TriBlk} & 51.65 & 30.50 & 47.89  \\
    \hline
    \hline
    \textsc{Lead3} & 40.42 & 17.62 & 36.67  \\
    \textsc{NN-SE}  & 40.81 & 17.91 & 37.03  \\
    \textsc{Seq2Seq} & 41.83 & 19.29 & 38.28  \\
    \textsc{SummaRuNNer}  & 41.84 & 19.31 & 38.31  \\
    \textsc{NeuSum} & 41.59 & 19.01 & 37.98 \\
    \textsc{HiBert} & 42.37 & 19.95 & 38.83 \\
    \hline
    \hline
    \textsc{BertSumExt} & 42.61 & 19.99 & 39.09  \\
    \textsc{BertSumExt}+\textsc{TriBlk} & 43.25 & 20.24 & 39.63  \\ 
    \hline
    \modelnameSEQ{} & 42.72$^{a}$ & 19.82 & 39.15  \\
    \modelnameCTX{} & \textbf{43.43}$^{AB}$ & \textbf{20.44}$^{AB}$ & \textbf{39.83}$^{AB}$  \\
    \bottomrule
    \end{tabular}
\end{table}
For \textsc{BertSumExt}-based methods, we observe that redundancy removal helps improve the \textsc{Rouge} score compared to \textsc{BertSumExt}. 
\textsc{TriBlk} has considerably better performances; \modelnameSEQ{} achieves better \textsc{Rouge}-1/L scores; \modelnameCTX{} significantly outperforms the other redundancy elimination methods. \footnote{The salience ranker of \modelnameCTX{} alone performs similarly to \textsc{BertSumExt}.} The performance differences of \modelnameCTX{} and \textsc{TriBlk} comes from $30.6\%$ summaries output by the two systems in the test set. In other cases, they agrees with each other. This shows that by \textit{adaptively} balancing salience and diversity, \modelnameCTX{} is superior to \textsc{TriBlk} when redundancy removal is promising. 



We also find that the sequence generation models, i.e., \textsc{NeuSum} and our \modelnameSEQ{}, do not have clear advantage over other models regardless of their encoder network structure (i.e., \textsc{Bert} or other neural architectures). For instance, \textsc{SummaRuNNer} and \textsc{Seq2Seq} models have the best performance among methods that are not based on \textsc{Bert}\footnote{The \textsc{Rouge} scores of \textsc{SummaRuNNer} are lower than \textsc{NeuSum} in \citet{zhou2018neural} because the results of \textsc{SummaRuNNer} are from the anonymized version CNN/DailyMail, which are not comparable with the results of \textsc{NeuSum} on the \textit{non-anonymized} version.}.
Our \modelnameSEQ{} perform similarly to \textsc{BertSumExt}.
\modelnameSEQ{} is inferior to \modelnameCTX{} due to its order-sensitive optimization objective. While \modelnameCTX{} learns to optimize towards all the possible ordering of the ground truth sentence set, \modelnameSEQ{} is optimized towards only one sequence of them. Another ordering of the same set will be penalized by \modelnameSEQ{} even though they have the same ROUGE score. Its significant worse $P@1$ (shown in Section \ref{subsec:model_ana}) also confirms this point. 

\begin{table}
    \caption{Full-length \textsc{Rouge} (RG) F1 (\%) evaluation on the NYT50 test set. a/b and A/B indicate significant \textsc{BertSumExt}/\textsc{BertSumExt}+\textsc{TriBlk} with $p<0.02$ and $p<0.0001$ respectively.}
    \centering
    \label{tab:nyt_result}  
    \small
    \begin{tabular}{l  p{0.9cm} p{0.9cm}  p{0.9cm} }
    \toprule
    Model & RG-1 & RG-2 & RG-L \\
    \toprule
    \textsc{Oracle} & 56.23 & 37.92 & 49.45 \\
    \textsc{Oracle}+\textsc{TriBlk} & 54.32 & 36.33 & 47.53 \\ 
    \hline
    \hline
    Lead3 & 38.20 & 19.29 & 30.49 \\
    \textsc{NN-SE} & 41.92 & 22.45 & 33.88 \\
    \textsc{Seq2Seq} & 44.45 & 24.72 & 36.20 \\
    \textsc{SummaRuNNer} & 44.70 & 24.87 & 36.44 \\ 
    \hline
    \hline
    \textsc{BertSumExt} & 45.46 & \textbf{25.53} & 37.17 \\
    \textsc{BertSumExt}+\textsc{TriBlk} & 44.90 & 24.87 & 36.63 \\ 
    \hline
    \modelnameSEQ{} & 45.15$^{B}$ & 25.14$^{B}$ & 36.79$^{B}$ \\
    \modelnameCTX{} & \textbf{45.54}$^{AB}$ & 25.52$^{B}$ & \textbf{37.22}$^{aB}$ \\
    \bottomrule
    \end{tabular}
\end{table}
\textbf{NYT50.} 
In contrast to CNN/DailyMail, we observe that \textsc{TriBlk} has harmed the performance of \textsc{BertSumExt} on NYT50.
In fact, as shown in Table \ref{tab:nyt_result}, applying \textsc{TriBlk} on \textsc{Oracle} also causes reduction in \textsc{Rouge}-1,2,L scores by 1.91, 1.59 and 1.92 absolute point respectively, which are much larger than those drops in CNN/DailyMail (0.94, 0.74 and 0.98).
This indicates that \textsc{TriBlk} filters out more sentences that have high \textsc{Rouge} gain on NYT50 than CNN/DailyMail, causing more drop of \textsc{Rouge}. It also shows that sentences in oracle summaries have more trigram overlap on NYT50 than CNN/DailyMail, which implies that redundancy removal on NYT50 may have limited gains and a simple unified rule (\textsc{TriBlk}) applying on all the documents could harm the performance. 

We also observe that \modelnameSEQ{} performs better than \textsc{BertSumExt}+\textsc{TriBlk} but worse than \textsc{BertSumExt}.
In contrast, \modelnameCTX{} achieves higher performance than \textsc{BertSumExt}+\textsc{TriBlk} and \modelnameSEQ{} by representing redundancy explicitly and controlling its effect dynamically. 
Since redundancy removal has a limited potential gain on NYT50, the predictions of \modelnameCTX{} differ from \textsc{BertSumExt} only in $10.1\%$ of the test set. 
However, these differences still lead to significant overall improvements. 

Note that the gain of \modelnameCTX{} comes only from redundancy removal, which takes effect from the second step of selecting sentences. The improvements can be larger when redundancy removal has higher potentials (e.g., CNN/DailyMail) and smaller on datasets (e.g., NYT50) with lower potentials. In either case, it does not harm the performances as the other methods, which shows that it is \textit{adaptive} and robust.

\subsection{Model Analysis}
\label{subsec:model_ana}
\begin{figure}
    \centering
    \includegraphics[width=0.45\textwidth]{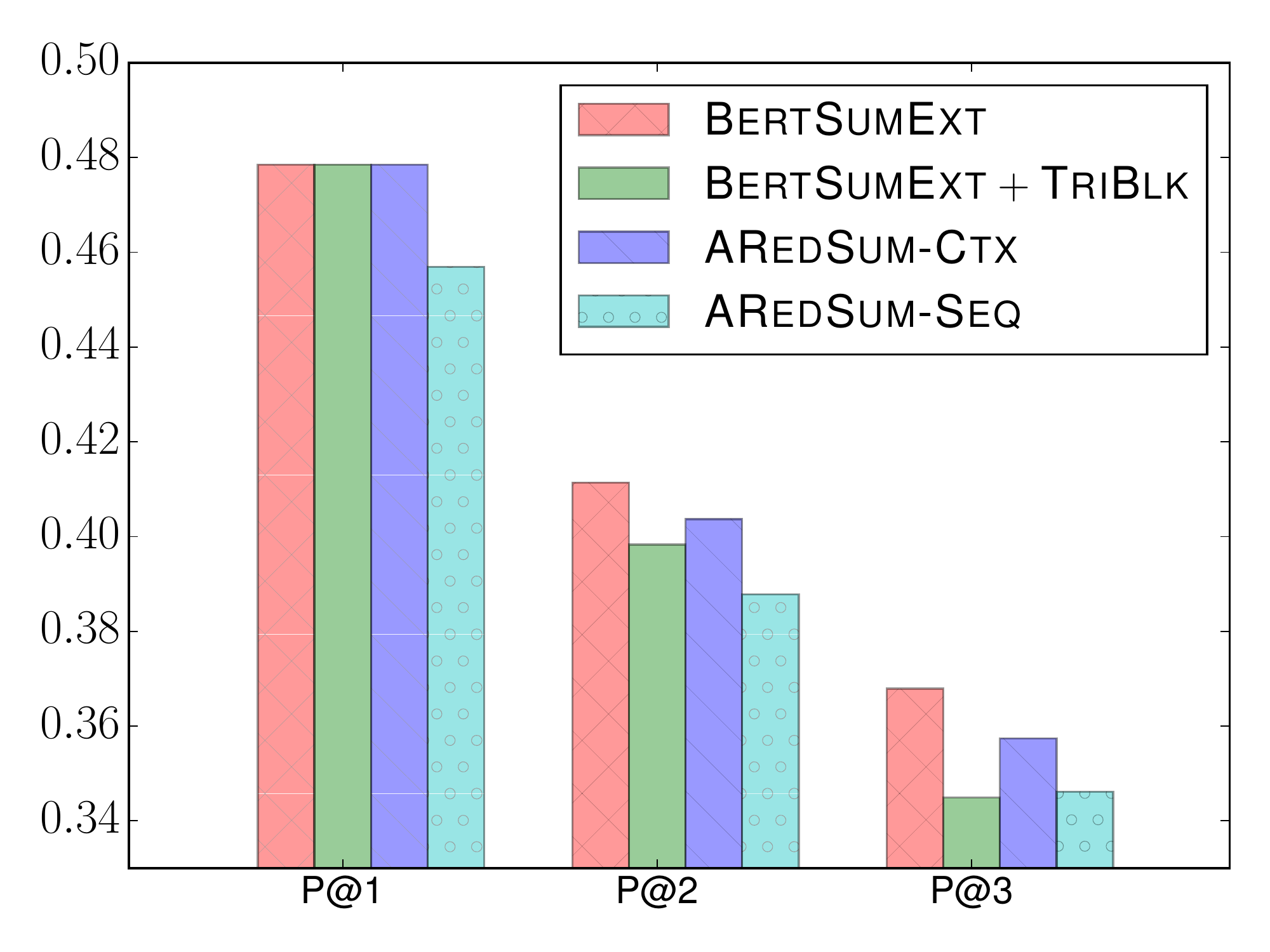} %
    \caption{The precision of the extracted sentences at step $k$ on CNN/DailyMail.}
\label{fig:cnndm_prec}
\end{figure}

\textbf{Precision at Each Step.}
Since the content of sentences with positive labels, i.e., $\hat{S}^*$, could vary from the original human-generated abstractive summaries $S^*$, 
models that have higher precision with respect to $\hat{S}^*$ do not necessarily yield better \textsc{Rouge} scores against $S^*$. 
Because \textsc{BertSumExt} is optimized towards $\hat{S}^*$ while \modelnameCTX{} and \modelnameSEQ{} aim to learn to select sentences with best \textsc{Rouge} gain against $S^*$, they behave differently in terms of \textsc{Rouge} and precision. 
Thus, we analyze how our model's selection at each $k$-th step affects the \textsc{Rouge} performance. 
We only present the precision on CNN/DailyMail in Figure \ref{fig:cnndm_prec} since similar trends are observed on NYT50. 
Note all the models except \modelnameSEQ{} have the same $P@1$ because initially, the model's selection is only based on salience. 
Filtering and demoting the selected sentences starts to take effect only after the second step.

As shown in the figure, \textsc{BertSumExt} has the best $P@1, P@2$ and $P@3$ among all, which is reasonable since $\hat{S}^*$ is the target which it is optimized to learn. 
When \textsc{TriBlk} is applied, $P@2$ and $P@3$ drop a lot while the \textsc{Rouge} scores are up (as in Table \ref{tab:cnndm_result}). 
It indicates that \textsc{TriBlk} could filter out some informative but redundant sentences during selection, which harms precision but improves \textsc{Rouge}.
In contrast, $P@2$ and $P@3$ of \modelnameCTX{} is between \textsc{BertSumExt} with and without \textsc{TriBlk}. 
Through learning towards \textsc{Rouge} gain given the previously extracted sentences, \modelnameCTX{} achieves the best \textsc{Rouge} scores with less harm to precision, which means that \modelnameCTX{} can better balance salience and redundancy.

\modelnameSEQ{} has a significantly lower $P@1$ than the others since its objective at the first step is to find the sentence with maximal \textsc{Rouge} gain, which is only one in $\hat{S}^*$. 
At steps 2 and 3, the disadvantage of \modelnameSEQ{} becomes smaller. It has similar $P@3$ to \textsc{BertSumExt}+\textsc{TriBlk}. The generated sequence of sentences cover a decent portion of $\hat{S}^*$, but it is still worse than the methods that do not use order-sensitive optimization objectives. 

\begin{figure}
    \centering
    \includegraphics[width=0.45\textwidth]{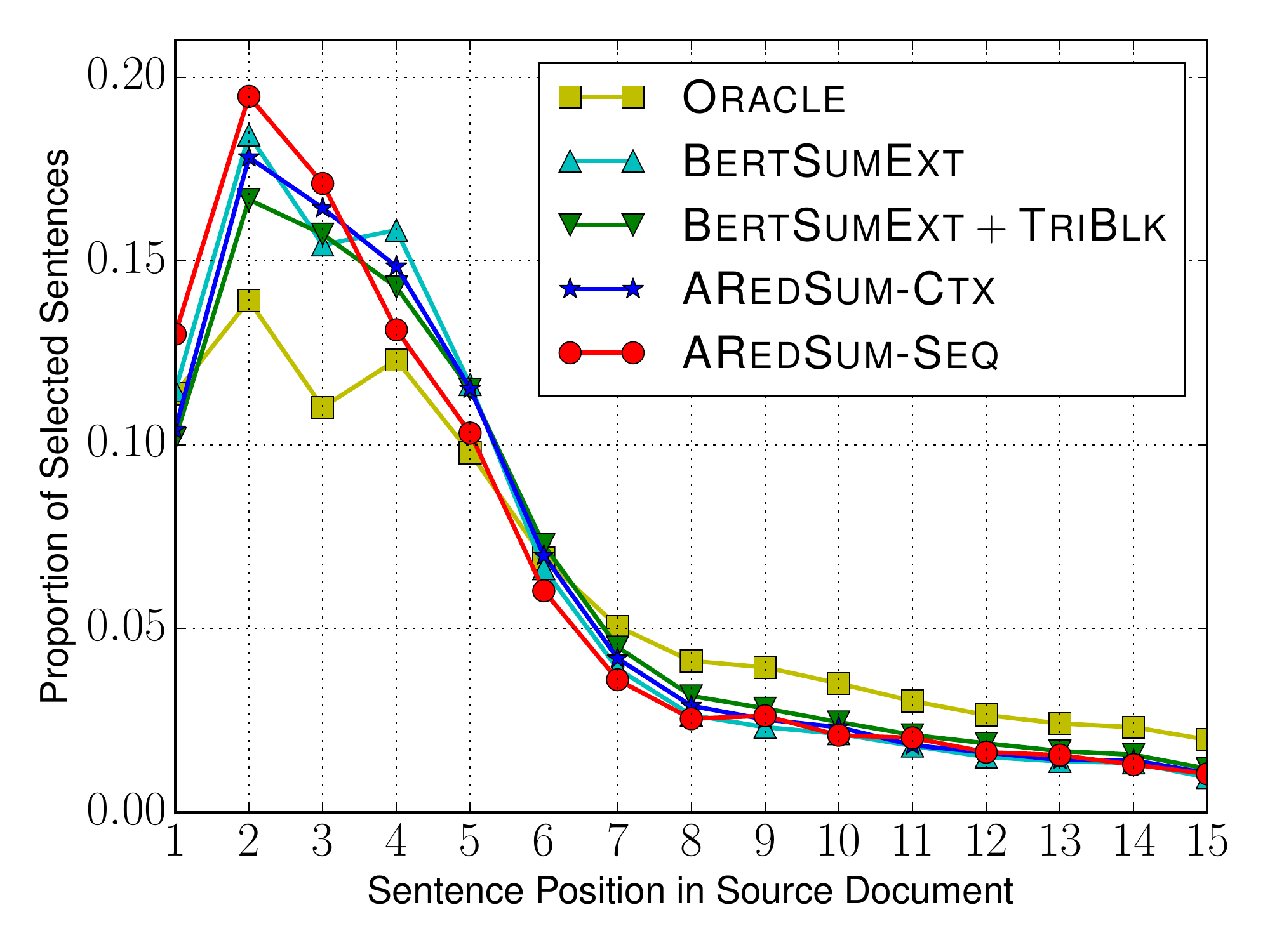} %
    \caption{The proportion of extracted sentences in terms of their position in the document.} 
\label{fig:cnndm_sent_pos}
\end{figure}

\textbf{Position of Selected Sentences.}
Figure \ref{fig:cnndm_sent_pos} shows the position of sentences extracted by different models and \textsc{Oracle} on CNN/DailyMail. A large portion of oracle sentences are the first 5 sentences, and all the models tend to extract the leading 5 sentences in the predicted summaries. The output of \modelnameSEQ{} concentrates more on the first 3 sentences, which differs from \textsc{Oracle} more than the other models. With \textsc{TriBlk}, \textsc{BertSumExt} selects sentences in later positions more. The position distribution of \modelnameCTX{} is between \textsc{BertSumExt} with and without \textsc{TriBlk}, which is similar to their precision distribution in Figure \ref{fig:cnndm_prec}. This indicates that \modelnameCTX{} seeks to find a smoother way to filter out sentences that are redundant but salient, and these sentences tend to be at earlier positions. 

\subsection{Human Evaluation}
We also conduct human evaluations to analyze how our best model compares against the best baseline model. On both datasets, we randomly sample 20 summaries constructed by the best baseline and our best model from the cases where their \textsc{Rouge-2} score difference is more than 0.05 points. Following \citet{zhou2018neural}, we asked two graduate student volunteers to rank the summaries extracted by different models from best to worst in terms of \textit{informativeness}, \textit{redundancy} and the \textit{overall} quality. We allowed ties in the analysis. Average ranks of the systems are shown in Table \ref{tab:human_eval}.

On CNN/DailyMail, \modelnameCTX{} ranks higher than \textsc{BertSumExt}+\textsc{TriBlk} in terms of each aspect. 
On NYT50, \modelnameCTX{} has a more compelling performance in terms of redundancy than informativeness. This is consistent with the fact that \textsc{BertSumExt} only focuses on learning salience and does not deal with redundancy during sentence selection. 
From both automatic and human evaluation of our best model and the best baseline, we can see that removing redundancy with our models is better than redundancy removal with heuristics and no redundancy removal. 

\begin{table}
    \caption{Average ranks of our best method and the best baseline on CNN/DailyMail and NYT50 in terms of informativeness (Info), redundancy (Rdnd) and the overall quality by human participants (the lower, the better). $*$ and $\dagger$ indicates significant improvements with $p<0.03$ and $p<0.0001$.}
    \centering
    \label{tab:human_eval}  
    \small
    \begin{tabular}{l c c c}
    \toprule
    \textbf{CNN/DailyMail} & Info & Rdnd & Overall \\
    \hline
    \textsc{BertSumExt} + \textsc{TriBlk} & 1.50 & 1.55 & 1.55 \\
    \modelnameCTX{} & 1.20 & 1.15$^*$ & 1.15$^*$ \\
    \toprule
    \textbf{NYT50} & Info & Rdnd & Overall \\
    \hline
    \textsc{BertSumExt} & 1.50 & 1.60 & 1.55 \\
    \modelnameCTX{} & 1.35 & 1.00$^\dagger$ & 1.35 \\
    \bottomrule
    \end{tabular}
\end{table}

%% file: conclusion.tex
\section{Conclusions}
Extending a state-of-the-art extractive summarization model, we propose \modelnameSEQ{} that jointly scores and selects sentences with a sequence generation model and \modelnameCTX{} that learns to balance salience and redundancy with a separate model. Experimental results show that \modelnameCTX{} outperforms  
\modelnameSEQ{} and all other strong baselines, which yields that redundancy reduction helps improve summary quality, and it is better to model the effect of redundancy explicitly than jointly with salience during sentence scoring.  

\section*{Acknowledgments}
This work was supported in part by the Center for Intelligent Information Retrieval. Any opinions, findings and conclusions or recommendations expressed in this material are those of the authors and do not necessarily reflect those of the sponsor.

%% file: main.bbl
\begin{thebibliography}{36}
\expandafter\ifx\csname natexlab\endcsname\relax\def\natexlab#1{#1}\fi

\bibitem[{Ai et~al.(2018)Ai, Bi, Guo, and Croft}]{ai2018learning}
Qingyao Ai, Keping Bi, Jiafeng Guo, and W~Bruce Croft. 2018.
\newblock Learning a deep listwise context model for ranking refinement.
\newblock In \emph{The 41st International ACM SIGIR Conference on Research \&
  Development in Information Retrieval}, pages 135--144. ACM.

\bibitem[{Cao et~al.(2015)Cao, Wei, Li, Li, Zhou, and Wang}]{cao2015learning}
Ziqiang Cao, Furu Wei, Sujian Li, Wenjie Li, Ming Zhou, and Houfeng Wang. 2015.
\newblock Learning summary prior representation for extractive summarization.
\newblock In \emph{Proceedings of the 53rd Annual Meeting of the Association
  for Computational Linguistics and the 7th International Joint Conference on
  Natural Language Processing (Volume 2: Short Papers)}, pages 829--833.

\bibitem[{Carbonell and Goldstein(1998)}]{carbonell1998use}
Jaime~G Carbonell and Jade Goldstein. 1998.
\newblock The use of mmr, diversity-based reranking for reordering documents
  and producing summaries.
\newblock In \emph{SIGIR}, volume~98, pages 335--336.

\bibitem[{Cheng and Lapata(2016)}]{cheng2016neural}
Jianpeng Cheng and Mirella Lapata. 2016.
\newblock Neural summarization by extracting sentences and words.
\newblock \emph{arXiv preprint arXiv:1603.07252}.

\bibitem[{Conroy et~al.(2004)Conroy, Schlesinger, Goldstein, and
  O’leary}]{conroy2004left}
John~M Conroy, Judith~D Schlesinger, Jade Goldstein, and Dianne~P O’leary.
  2004.
\newblock Left-brain/right-brain multi-document summarization.
\newblock In \emph{Proceedings of the Document Understanding Conference (DUC
  2004)}.

\bibitem[{Devlin et~al.(2018)Devlin, Chang, Lee, and
  Toutanova}]{devlin2018bert}
Jacob Devlin, Ming-Wei Chang, Kenton Lee, and Kristina Toutanova. 2018.
\newblock Bert: Pre-training of deep bidirectional transformers for language
  understanding.
\newblock \emph{arXiv preprint arXiv:1810.04805}.

\bibitem[{Dong et~al.(2018)Dong, Shen, Crawford, van Hoof, and
  Cheung}]{dong2018banditsum}
Yue Dong, Yikang Shen, Eric Crawford, Herke van Hoof, and Jackie Chi~Kit
  Cheung. 2018.
\newblock Banditsum: Extractive summarization as a contextual bandit.
\newblock \emph{arXiv preprint arXiv:1809.09672}.

\bibitem[{Durrett et~al.(2016)Durrett, Berg-Kirkpatrick, and
  Klein}]{durrett2016learning}
Greg Durrett, Taylor Berg-Kirkpatrick, and Dan Klein. 2016.
\newblock Learning-based single-document summarization with compression and
  anaphoricity constraints.
\newblock \emph{arXiv preprint arXiv:1603.08887}.

\bibitem[{Erkan and Radev(2004)}]{erkan2004lexrank}
G{\"u}nes Erkan and Dragomir~R Radev. 2004.
\newblock Lexrank: Graph-based lexical centrality as salience in text
  summarization.
\newblock \emph{Journal of artificial intelligence research}, 22:457--479.

\bibitem[{Galley(2006)}]{galley2006skip}
Michel Galley. 2006.
\newblock A skip-chain conditional random field for ranking meeting utterances
  by importance.
\newblock In \emph{Proceedings of the 2006 Conference on Empirical Methods in
  Natural Language Processing}, pages 364--372. Association for Computational
  Linguistics.

\bibitem[{Hermann et~al.(2015)Hermann, Kocisky, Grefenstette, Espeholt, Kay,
  Suleyman, and Blunsom}]{hermann2015teaching}
Karl~Moritz Hermann, Tomas Kocisky, Edward Grefenstette, Lasse Espeholt, Will
  Kay, Mustafa Suleyman, and Phil Blunsom. 2015.
\newblock Teaching machines to read and comprehend.
\newblock In \emph{Advances in neural information processing systems}, pages
  1693--1701.

\bibitem[{Kedzie et~al.(2018)Kedzie, McKeown, and
  Daume~III}]{kedzie2018content}
Chris Kedzie, Kathleen McKeown, and Hal Daume~III. 2018.
\newblock Content selection in deep learning models of summarization.
\newblock \emph{arXiv preprint arXiv:1810.12343}.

\bibitem[{Kupiec et~al.(1999)Kupiec, Pedersen, and Chen}]{kupiec1999trainable}
Julian Kupiec, Jan Pedersen, and Francine Chen. 1999.
\newblock A trainable document summarizer.
\newblock \emph{Advances in Automatic Summarization}, pages 55--60.

\bibitem[{Lin(2004)}]{lin2004rouge}
Chin-Yew Lin. 2004.
\newblock Rouge: A package for automatic evaluation of summaries.
\newblock In \emph{Text summarization branches out}, pages 74--81.

\bibitem[{Lin and Bilmes(2011)}]{lin2011class}
Hui Lin and Jeff Bilmes. 2011.
\newblock A class of submodular functions for document summarization.
\newblock In \emph{Proceedings of the 49th Annual Meeting of the Association
  for Computational Linguistics: Human Language Technologies-Volume 1}, pages
  510--520. Association for Computational Linguistics.

\bibitem[{Liu et~al.(2009)}]{liu2009learning}
Tie-Yan Liu et~al. 2009.
\newblock Learning to rank for information retrieval.
\newblock \emph{Foundations and Trends{\textregistered} in Information
  Retrieval}, 3(3):225--331.

\bibitem[{Liu and Lapata(2019)}]{liu2019text}
Yang Liu and Mirella Lapata. 2019.
\newblock Text summarization with pretrained encoders.
\newblock \emph{Proceedings of the 2019 Conference on Empirical Methods in
  Natural Language Processing and the 9th International Joint Conference on
  Natural Language Processing (EMNLP-IJCNLP)}.

\bibitem[{McDonald(2007)}]{mcdonald2007study}
Ryan McDonald. 2007.
\newblock A study of global inference algorithms in multi-document
  summarization.
\newblock In \emph{European Conference on Information Retrieval}, pages
  557--564. Springer.

\bibitem[{Mihalcea and Tarau(2004)}]{mihalcea2004textrank}
Rada Mihalcea and Paul Tarau. 2004.
\newblock Textrank: Bringing order into text.
\newblock In \emph{Proceedings of the 2004 conference on empirical methods in
  natural language processing}, pages 404--411.

\bibitem[{Nallapati et~al.(2017)Nallapati, Zhai, and
  Zhou}]{nallapati2017summarunner}
Ramesh Nallapati, Feifei Zhai, and Bowen Zhou. 2017.
\newblock Summarunner: A recurrent neural network based sequence model for
  extractive summarization of documents.
\newblock In \emph{Thirty-First AAAI Conference on Artificial Intelligence}.

\bibitem[{Narayan et~al.(2018)Narayan, Cohen, and Lapata}]{narayan2018ranking}
Shashi Narayan, Shay~B Cohen, and Mirella Lapata. 2018.
\newblock Ranking sentences for extractive summarization with reinforcement
  learning.
\newblock \emph{arXiv preprint arXiv:1802.08636}.

\bibitem[{Nenkova et~al.(2006)Nenkova, Vanderwende, and
  McKeown}]{nenkova2006compositional}
Ani Nenkova, Lucy Vanderwende, and Kathleen McKeown. 2006.
\newblock A compositional context sensitive multi-document summarizer:
  exploring the factors that influence summarization.
\newblock In \emph{Proceedings of the 29th annual international ACM SIGIR
  conference on Research and development in information retrieval}, pages
  573--580. ACM.

\bibitem[{Osborne(2002)}]{osborne2002using}
Miles Osborne. 2002.
\newblock Using maximum entropy for sentence extraction.
\newblock In \emph{Proceedings of the ACL-02 Workshop on Automatic
  Summarization-Volume 4}, pages 1--8. Association for Computational
  Linguistics.

\bibitem[{Paulus et~al.(2017)Paulus, Xiong, and Socher}]{paulus2017deep}
Romain Paulus, Caiming Xiong, and Richard Socher. 2017.
\newblock A deep reinforced model for abstractive summarization.
\newblock \emph{arXiv preprint arXiv:1705.04304}.

\bibitem[{Ren et~al.(2017)Ren, Chen, Ren, Wei, Ma, and
  de~Rijke}]{ren2017leveraging}
Pengjie Ren, Zhumin Chen, Zhaochun Ren, Furu Wei, Jun Ma, and Maarten de~Rijke.
  2017.
\newblock Leveraging contextual sentence relations for extractive summarization
  using a neural attention model.
\newblock In \emph{Proceedings of the 40th International ACM SIGIR Conference
  on Research and Development in Information Retrieval}, pages 95--104.

\bibitem[{Ren et~al.(2016)Ren, Wei, Chen, Ma, and
  Zhou}]{ren-etal-2016-redundancy}
Pengjie Ren, Furu Wei, Zhumin Chen, Jun Ma, and Ming Zhou. 2016.
\newblock A redundancy-aware sentence regression framework for extractive
  summarization.
\newblock In \emph{Proceedings of {COLING} 2016, the 26th International
  Conference on Computational Linguistics: Technical Papers}, pages 33--43,
  Osaka, Japan.

\bibitem[{Sandhaus(2008)}]{sandhaus2008new}
Evan Sandhaus. 2008.
\newblock The new york times annotated corpus.
\newblock \emph{Linguistic Data Consortium, Philadelphia}, 6(12):e26752.

\bibitem[{See et~al.(2017)See, Liu, and Manning}]{see2017get}
Abigail See, Peter~J Liu, and Christopher~D Manning. 2017.
\newblock Get to the point: Summarization with pointer-generator networks.
\newblock \emph{arXiv preprint arXiv:1704.04368}.

\bibitem[{Vaswani et~al.(2017)Vaswani, Shazeer, Parmar, Uszkoreit, Jones,
  Gomez, Kaiser, and Polosukhin}]{vaswani2017attention}
Ashish Vaswani, Noam Shazeer, Niki Parmar, Jakob Uszkoreit, Llion Jones,
  Aidan~N Gomez, {\L}ukasz Kaiser, and Illia Polosukhin. 2017.
\newblock Attention is all you need.
\newblock In \emph{Advances in neural information processing systems}, pages
  5998--6008.

\bibitem[{Wan and Yang(2006)}]{wan2006improved}
Xiaojun Wan and Jianwu Yang. 2006.
\newblock Improved affinity graph based multi-document summarization.
\newblock In \emph{Proceedings of the human language technology conference of
  the NAACL, Companion volume: Short papers}, pages 181--184. Association for
  Computational Linguistics.

\bibitem[{Wan and Yang(2008)}]{wan2008multi}
Xiaojun Wan and Jianwu Yang. 2008.
\newblock Multi-document summarization using cluster-based link analysis.
\newblock In \emph{Proceedings of the 31st annual international ACM SIGIR
  conference on Research and development in information retrieval}, pages
  299--306. ACM.

\bibitem[{Wang et~al.(2020)Wang, Liu, Zheng, Qiu, and
  Huang}]{wang2020heterogeneous}
Danqing Wang, Pengfei Liu, Yining Zheng, Xipeng Qiu, and Xuanjing Huang. 2020.
\newblock Heterogeneous graph neural networks for extractive document
  summarization.
\newblock \emph{arXiv preprint arXiv:2004.12393}.

\bibitem[{Zhang et~al.(2019)Zhang, Wei, and Zhou}]{zhang-etal-2019-hibert}
Xingxing Zhang, Furu Wei, and Ming Zhou. 2019.
\newblock \href {https://doi.org/10.18653/v1/P19-1499} {{HIBERT}: Document
  level pre-training of hierarchical bidirectional transformers for document
  summarization}.
\newblock In \emph{Proceedings of the 57th Annual Meeting of the Association
  for Computational Linguistics}, pages 5059--5069.

\bibitem[{Zhong et~al.(2019)Zhong, Liu, Wang, Qiu, and
  Huang}]{zhong2019searching}
Ming Zhong, Pengfei Liu, Danqing Wang, Xipeng Qiu, and Xuanjing Huang. 2019.
\newblock Searching for effective neural extractive summarization: What works
  and what's next.
\newblock \emph{arXiv preprint arXiv:1907.03491}.

\bibitem[{Zhou et~al.(2020)Zhou, Wei, and Zhou}]{zhou2020level}
Qingyu Zhou, Furu Wei, and Ming Zhou. 2020.
\newblock At which level should we extract? an empirical study on extractive
  document summarization.
\newblock \emph{arXiv preprint arXiv:2004.02664}.

\bibitem[{Zhou et~al.(2018)Zhou, Yang, Wei, Huang, Zhou, and
  Zhao}]{zhou2018neural}
Qingyu Zhou, Nan Yang, Furu Wei, Shaohan Huang, Ming Zhou, and Tiejun Zhao.
  2018.
\newblock Neural document summarization by jointly learning to score and select
  sentences.
\newblock \emph{arXiv preprint arXiv:1807.02305}.

\end{thebibliography}
